\documentclass[10pt,twocolumn,letterpaper]{article}

 \usepackage[pagenumbers]{cvpr} 

\usepackage[dvipsnames]{xcolor}

\usepackage{algorithm2e}
\usepackage{amsmath}
\usepackage{placeins}

\definecolor{cvprblue}{rgb}{0.21,0.49,0.74}
\usepackage[pagebackref,breaklinks,colorlinks,citecolor=cvprblue]{hyperref}

\def\paperID{4} 
\def\confName{CVPR}
\def\confYear{2024}

\title{Reinforcement Learning with Generative Models for Compact Support Sets}

\author{Nico Schiavone\\
University of Alberta\\
{\tt\small nschiavo@ualberta.ca}
\and
Xingyu Li\\
University of Alberta\\
{\tt\small xingyu@ualberta.ca}
}

\begin{document}
\maketitle
\begin{abstract}
Foundation models contain a wealth of information from their vast number of training samples. However, most prior arts fail to extract this information in a precise and efficient way for small sample sizes. In this work, we propose a framework utilizing reinforcement learning as a control for foundation models, allowing for the granular generation of small, focused synthetic support sets to augment the performance of neural network models on real data classification tasks. We first allow a reinforcement learning agent access to a novel context based dictionary; the agent then uses this dictionary with a novel prompt structure to form and optimize prompts as inputs to generative models, receiving feedback based on a reward function combining the change in validation accuracy and entropy. A support set is formed this way over several exploration steps. Our framework produced excellent results, increasing classification accuracy by significant margins for no additional labelling or data cost.
\end{abstract}    
\section{Introduction}
\label{sec:intro}

Deep learning~\cite{deeplearning} is one of the most popular and successful methods for any task where a large dataset can be procured, including fundamental computer vision tasks like classification. However, large, well-balanced, well-labelled datasets are often difficult and prohibitively expensive to acquire. Consequently, much of contemporary image classification utilizes a high quality source dataset and support sets with highly relevant data to the target task. The generation of such support sets has been a focus of contemporary research, and recently utilizes the output of the unprecedented success of large pretrained generative models like Stable Diffusion~\cite{stablediffusion}. The advancements in generative models have led to the rise of synthetic datasets, where images are generated in large scale according to the target task and used in place of a real training dataset, yielding excellent results~\cite{thinair, issyntheticready, nakamura2023pretraining}.

Despite these advancements, the body of research relating to synthetic datasets remains primarily focused on large-batch image synthesis. In this way, any issues caused by the unpredictable behaviour of modern generative models can easily be smoothed out. However, this results in the majority of successful applications requiring tens of thousands of images generated for a single task~\cite{nakamura2023pretraining, issyntheticready}, which is inefficient in time and cost.

The goal of creating specific, highly focused support sets composed of several hundred images rather than several thousand is currently an open problem at the forefront of generative computer vision research. Consequently, it raises the question of if synthetic data can supplement real data, making up a very small portion of the overall dataset to shore up specific weaknesses, or whether synthetic data must make up a significant amount of the dataset if it is to be used at all.

Reinforcement learning~\cite{ppo} is a popular control scheme that has an agent learn the optimal behaviour given an environment and a reward for desirable interactions. Recent studies have found reinforcement learning effective at writing and re-writing prompts~\cite{kong2024prewrite, deng2022rlprompt}, but the use of reinforcement learning to guide the evolution of prompts has yet to be explored. Reinforcement learning is an excellent framework for imposing specific learned behaviours upon the resulting agent, and we posit that combining reinforcement learning with pretrained generative models will impart that much-needed specificity on the synthesized images, resulting in significant performance gains for a relatively small number of synthetic images.

In this work, we introduce a framework utilizing reinforcement learning as a control for large generative models to synthesize precise support sets, intended to bolster the lacking aspects of real datasets without overwriting them for increased model performance at no extra data or labelling costs. To accomplish this, we utilize a dictionary based on the features of the original training dataset, and allow a reinforcement learning agent to learn the optimal structures and word choice to generate high quality, specific prompts for Stable Diffusion. The controlled output of Stable Diffusion is then used to supplement the existing training data for a neural network model, and the performance of this model on a validation set is given as feedback to the agent. In this way, the framework allows Stable Diffusion to act as an extension of the reinforcement learning agent, acting directly to improve the performance of the model by tweaking the prompts that make up the support set. We evaluate this framework on several datasets, including CIFAR-10~\cite{cifar}, and Tiny-ImageNet~\cite{tinyimagenet}, showing free improvements on neural networks of $\sim$1\% for less than 500 total images in the support set. 

The main contributions for this work are:
\begin{itemize}
    \item A novel framework combining reinforcement learning and large pretrained generative models for the construction of small, focused, and effective synthetic support sets.
    \item A new reward scheme that facilitates a better interaction between reinforcement learning and classification.
\end{itemize}
\section{Related Work}
\label{sec:relwork}

\subsection{Reinforcement Learning}
Reinforcement learning~\cite{ppo} defines an agent and an environment with rules on how they can interact. The agent receives rewards based on how their actions affect the environment, with one of several reward schemes. The rewards inform the optimal behaviour of the agent, and thus the desirable properties of the end model. Popular reward schemes include exploration-based, which incentivizes exploring the action space, and goal-based, which explores to achieve set goals.

Past works have attempted to use reinforcement learning directly in classification algorithms, but this generally yields lacklustre results for the amount of effort and training time required~\cite{hafiz2020image}. This is due to the long convergence time of conventional reinforcement learning algorithms, and the relative ease of using simple deep learning models when a well-labelled dataset is available, rather than optimizing the loss with an agent. In our framework, we circumvent this issue by using a deep learning model for classification and optimizing it by altering the training set, rather than directly making the predictions using the agent.

\subsection{Generative Models}

Generative models have shown unprecedented success in many tasks in natural language processing and computer vision~\cite{gpt, stablediffusion}. Such models are often trained on datasets with in excess of one billion images, which stores a large wealth of knowledge that can be accessed through their generation capabilities~\cite{gpt}. These generative models have been widely used in contemporary research for image synthesis, such as augmentation of existing samples to artificially simulate a larger dataset~\cite{aug1, aug2}. Replacing the dataset entirely with synthetic images is also a topic of interest, with excellent preliminary results despite no real data~\cite{thinair}. Finally, the generation of large support sets to supplement real data has also been explored, but this mainly utilizes synthesis over a large scale to shore up the weaknesses of a dataset~\cite{nakamura2023pretraining}.

Contemporary generative models usually require text prompts to guide their behaviour. General prompting is successful in simple tasks, such as guided image synthesis, but complex and specific prompts often lead to unexpected results. This leads to an area of research known as prompt engineering, which is the focus of much of the recent literature in the topic of large models~\cite{chen2023unleashing}. Common approaches generally utilize a fixed set of prompts that have been carefully engineered to produce certain results; in our framework, we allow the prompts to evolve naturally from a general structure to their optimal state using reinforcement learning to choose the subjects and the model performance as feedback.
\section{Methods}
\label{sec:methods}

\subsection{Problem Formulation}
Initially, there is a well-labelled dataset $\mathcal{D}$, consisting of $N$ training samples, and a synthetic support set $\mathcal{S}$, consisting of $k*m$ samples, where $k$ is the current step number, and $m$ is the number of samples generated per step. In this work, we impose an extra limit $N_{\textrm{syn}}$ on the number of samples in $\mathcal{S}$. There is also a validation set $\mathcal{V}$, and a test set $\mathcal{T}$.

\begin{figure}[t]
    \centering
    \includegraphics[width=\columnwidth]{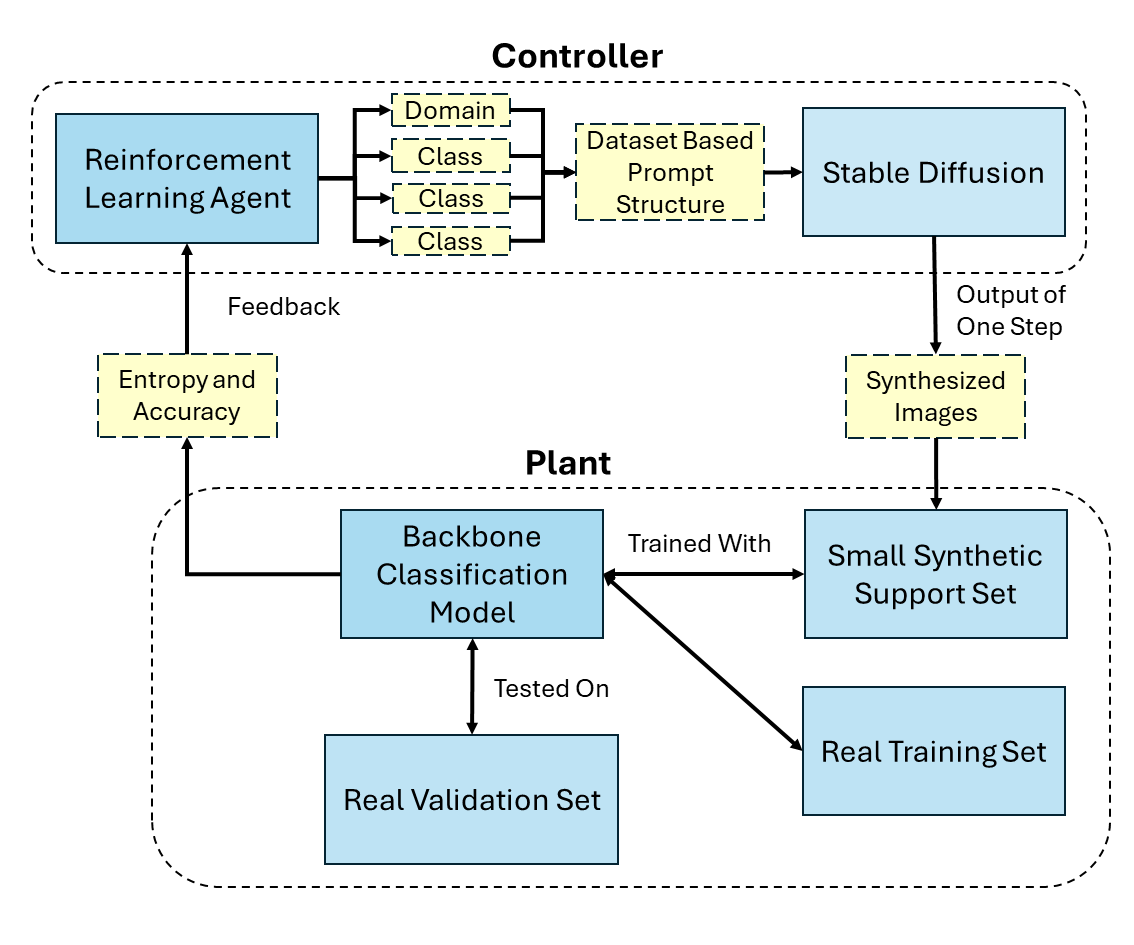}
    \caption{Overall framework}
    \label{fig:framework}
\end{figure}

Our goal in this study is to train a reinforcement learning agent $\mathcal{A}$ to optimally control a pretrained generative model, such as Stable Diffusion, to optimally populate $\mathcal{S}$ with at most $N_{\textrm{syn}}$ synthetic images, where $N_{\textrm{syn}} << N$. As shown in Fig. \ref{fig:framework}, in each step, the agent forms a prompt, feeds it to Stable Diffusion, and the resulting images are added to $\mathcal{S}$. The resulting dataset $\mathcal{D} + \mathcal{S}$ is used to train a model $\mathcal{M}$ , and its performance on $\mathcal{V}$ is passed back to $\mathcal{A}$ as feedback. This continues until a total of $N_{\textrm{syn}}$ images are contained within $\mathcal{S}$, at which point the exploration thread terminates. When all exploration threads within the preset exploration budget are explored, the resulting framework is tested on the test set $\mathcal{T}$ yielding the final performance.

\begin{figure}[t]
    \centering
    \includegraphics[width=0.8\columnwidth]{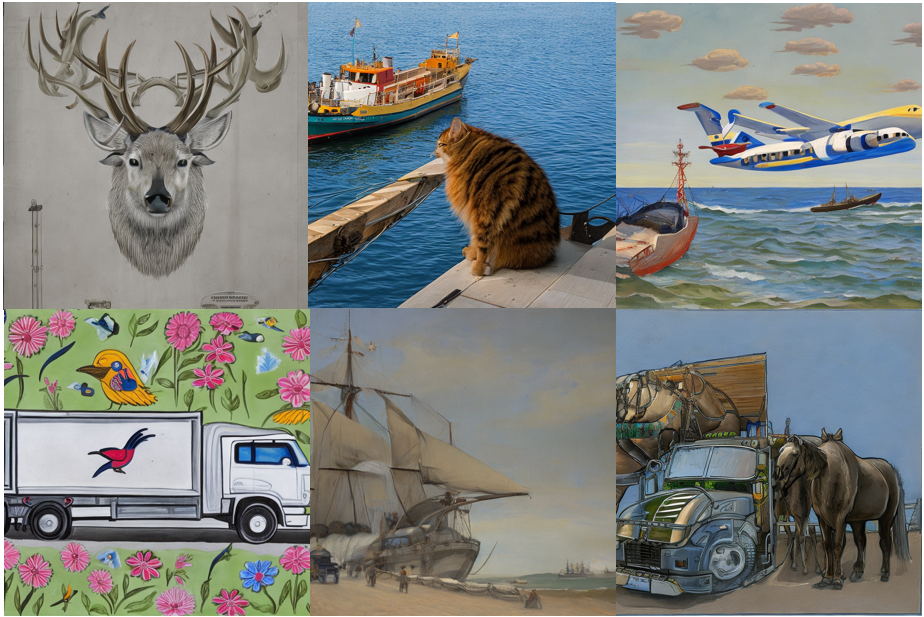}
    \caption{Images generated using our framework using CIFAR-10~\cite{cifar} labels.}
    \label{fig:images}
\end{figure}

\subsection{Image Synthesis}
For image synthesis, we are using Stable Diffusion~\cite{stablediffusion}, a successful text-to-image model that is trained on billions of text-image pairs.
Stable Diffusion has already been used to great effect in contemporary works when the aim is to replace a real dataset~\cite{stablerep, thinair}, and to augment existing samples~\cite{aug1, aug2}, but with comparatively fewer works focusing on consistently generating small, effective support sets.

\subsection{Controlling the Synthesis with RL}

Reinforcement learning (RL) defines an agent and an environment, and gives a set of actions that the agent can take to affect the environment. In our framework, we take a classification model and its training dataset as the \underline{Environment}. The reinforcement learning agent adaptively selects text prompts for the generative model towards image synthesis, which supplements the training set for classification performance improvement. The agent then receives feedback based on the change in the model's performance, which is taken as the \underline{State} in our reinforcement framework. In this study, we adopt the policy-based method for agent optimization, building a policy $\pi : s \longrightarrow a$ that maps states to optimal actions~\cite{ppo}. The specific objective function is:
\begin{equation}
    \label{eq:ppo}
    L(\theta) = \\ \hat{\mathbb{E}}[\textrm{min}(r_t(\theta) \hat{A}_t, \textrm{clip}(r_t(\theta), 1 - \epsilon, 1 + \epsilon)\hat{A}_t)].
\end{equation}
where $r_t = \frac{\pi_{\theta}(a_t | s_t)}{\pi_{\theta_\textrm{old}}(a_t | s_t)}$ is the probability ratio, $\hat{A}_t$ is the estimator of the advantage function at step $t$, and $\epsilon$ is a small value.

\underline{Action space}: Our framework allows the reinforcement learning agent to interact with Stable Diffusion by forming prompts. Prompts of unlimited length are subject to unmanageable time complexity, so we utilize a set dictionary based on the dataset. We formulate the interaction with a basic sentence structure with enough expression to accurately place the image, and pose the following format: "\textit{A \{domain\} of a \{class\}, \{class\}, and \{class\}}". Domains include photographs, digital artwork, paintings, mosaics, and other clear representations of the class. Next, three class names are chosen from the list of classes in the dataset. We notice that Stable Diffusion usually puts more attention on the first "class" term and generates the corresponding theme in the resulting image. Thus, our prompt design allows the agent to position the generated images at the boundaries between classes, which is where new images are most effective for improving classification performance~\cite{hardneg}. This is in contrast to traditional prompting methods, where the prompt describes the primary subject of interest with qualifiers for other subjects. We instead follow contemporary diversity research, prioritizing brevity and maximal control~\cite{shipard2023diversity}. 

The benefits of our approach are that single-class representative samples can be easily generated as follows: "\textit{A \{domain\} of a {car}, {car}, and {car}}", which has the added benefit of including more representative features from the chosen class due to the repetition. Multi-class samples can be equally easily generated by including two or three different class names, and the significance of each class can be altered by changing the order the classes appear in. In this way, our method allows the agent a yet unseen amount of control over the output of Stable Diffusion, resulting in significantly improved precision.

\underline{Reward function:} The agent's desired behaviour is to increase the accuracy of the classification model as much as possible with limited image synthesis. In our framework, we use a combined reward function, utilizing the validation set accuracy and the entropy to bias our model towards high, balanced accuracy. Under the assumption of a well-labelled training dataset, the former (i.e. classification accuracy on validate set) offers the most unfiltered access to the state changes in the model's performance. It is noteworthy that different from previous works utilizing reinforcement learning for classification, the accuracy alone is used, the addition of entropy in our reward allows the framework to simultaneously reward the improvement of weak classes, which improves the overall model performance on underrepresented classes. 

The formulation of our reward function is shown in Eq.~\ref{eq:reward}, where the entropy under a state $s$ can be calculated following Eq.~\ref{eq:entropy}. 
\begin{equation}
    \label{eq:reward}
    r(s, s') = \Delta\textrm{Acc}(s \rightarrow s') - \Delta\sigma_{\textbf{entropy}}(s \rightarrow s') ,
\end{equation}
\begin{equation}
    \label{eq:entropy}
    \sigma_{\textbf{entropy}}(\textbf{x}, \mathcal{M}) = - \Sigma_{i=1}^k p_\mathcal{M}(y_i | \textbf{x}) \log{p_\mathcal{M}(y_i | \textbf{x})},
\end{equation}
where $s'$ is the state after performing action $a$, and $s$ is the state before performing action $a$, and $p_\mathcal{M}(\hat{y} | \textbf{x})$ represents the class probability of sample $\textbf{x}$ under model $\mathcal{M}$.

\subsection{Full Algorithm}
One training step for the agent $\mathcal{A}$ consists of the following processes, in order:

\begin{enumerate}
    \item $\mathcal{A}$ chooses a \textit{domain} and three \textit{classes} in the prompt to represent the generated images.
    \item $m$ images are generated following the prompt, which are added to $\mathcal{S}$.
    \item $\mathcal{M}$ is trained on $\mathcal{D} + \mathcal{S}$, and tested again on $\mathcal{V}$, reporting the accuracy and entropy of the predictions.
    \item The reward $r(s, s')$ is given back to the agent. If $k=1$, then the pretrained statistics are used in place of the data from the previous state $s$.
\end{enumerate}

This sequence is optimized using Proximal-Policy-Optimization~\cite{ppo} to find the optimal set of $N_{\textrm{syn}}$ synthetic samples contained in $\mathcal{S}$. After the training process is completed, the algorithm has found the optimal prompts for to generate the optimal support set, and runs a final time without feedback to form $\mathcal{S}$, the desired support set.

\section{Results \& Discussion}
\label{sec:results}


\begin{table}[t]
\small
    \centering
    \begin{tabular}{|c|ccc|}
    \hline
                    & \multicolumn{1}{c|}{Pretrained} & \multicolumn{1}{c|}{Rand Syn} & Ours          \\ \hline
    ResNet-18       & \multicolumn{1}{c|}{92.0}       & \multicolumn{1}{c|}{92.3}             & \textbf{92.7} \\
    ResNet-50       & \multicolumn{1}{c|}{93.9}       & \multicolumn{1}{c|}{94.2}             & \textbf{94.5} \\
    VGG-16          & \multicolumn{1}{c|}{93.9}       & \multicolumn{1}{c|}{94.1}             & \textbf{94.9} \\
    ShuffleNetV2    & \multicolumn{1}{c|}{93.3}       & \multicolumn{1}{c|}{93.6}             & \textbf{94.1} \\
    EfficientNetV2S & \multicolumn{1}{c|}{94.1}       & \multicolumn{1}{c|}{94.3}             & \textbf{95.2} \\ \hline
    \end{tabular}
    \caption{Classification accuracy (\%) on CIFAR-10~\cite{cifar}.}
    \label{tab:res1}
\end{table}

\begin{table}[t]
\small
    \centering
    \begin{tabular}{|c|ccc|}
    \hline
                           & \multicolumn{1}{c|}{Pretrained} & \multicolumn{1}{c|}{Rand Syn} & Ours          \\ \hline
    ResNet-18              & \multicolumn{1}{c|}{54.3}       & \multicolumn{1}{c|}{54.4}             & \textbf{54.7} \\
    ResNet-50              & \multicolumn{1}{c|}{71.1}       & \multicolumn{1}{c|}{71.1}             & \textbf{71.5} \\
    VGG-16                 & \multicolumn{1}{c|}{63.2}       & \multicolumn{1}{c|}{63.4}             & \textbf{63.9} \\
    ShuffleNetV2           & \multicolumn{1}{c|}{48.6}       & \multicolumn{1}{c|}{48.6}             & \textbf{48.8} \\
    EfficientNetV2S        & \multicolumn{1}{c|}{69.9}       & \multicolumn{1}{c|}{70.0}             & \textbf{70.4} \\ \hline
    \end{tabular}
    \caption{Classification accuracy (\%) on Tiny ImageNet~\cite{tinyimagenet}.}
    \label{tab:res2}
\end{table}


\subsection{Datasets}
We evaluate our framework on two popular natural image datasets, CIFAR-10~\cite{cifar} and Tiny ImageNet~\cite{tinyimagenet}. We chose these datasets due to computational reasons -- the action space complexity scales as $n^3$, where $n$ is the number of classes in the dataset. Tiny ImageNet is a 200 class balanced dataset of 100 000 64x64 coloured images, and CIFAR-10 is a 10 class balanced dataset of 60 000 32x32 coloured images. In each case, we split the datasets using an 80:10:10 ratio of train:validation:test.

\subsection{Experimental Protocol}
We follow the setup laid out in Section~\ref{sec:methods}. For both datasets, we use a domain dictionary of \{"photograph", "painting", "still-life", "image", "digital image"\} and a class dictionary composed of each class name once. In experiments, we select $k=10$ to generate 10 images per step and our algorithm will run until a maximum of $N_{\textrm{syn}} = 400$ images.

Various models, including ResNet18, ResNet50~\cite{resnet}, ShuffleNetV2~\cite{efficientnetv2}, VGG-16~\cite{vgg}, and EfficientNetV2~\cite{zhang2017shufflenet}, are evaluated in our experiments. We compare the results of our framework against vanilla trained models and the models trained with random synthetic images in equal number. 
The 'Random Synthesis' setting adds to the training set 400 images synthesized by selecting random classes to fill the blanks in the prompt, and our method uses the full reinforcement learning framework.

\subsection{Main Results and Discussion}

The results of applying our framework are reported in Tables~\ref{tab:res1} and~\ref{tab:res2}. In addition, example images generated off of the CIFAR-10 dataset are demonstrated in Fig.~\ref{fig:images}. From these results, we can see that our framework is superior to random synthesis for small-batch support set synthesis, increasing the accuracy by as much as 0.9\% over the random synthesis method, and 1.1\% over the baseline model. Notably, for two backbones on Tiny ImageNet, random synthesis fails to improve the performance of the model by $>0.1\%$, while our framework increases the accuracy by $\sim$0.2\%. In addition, our method adds only 0.33\% extra images for CIFAR-10, and 0.2\% for Tiny-ImageNet. 

Our experimental results show that the proposed framework has a high performance gain relative to the number of samples synthesized, a characteristic not seen in prior arts. We attribute this gain to the fine control that our designed reinforcement learning agent gives over the output of the large pretrained model, and the effectiveness of the feedback given back to the agent.

Our framework currently requires some amount of information about the target dataset in order to work: class names, and a rough domain. This could be bypassed by forming the dictionary using an image-to-text encoder on representative samples after clustering by an unsupervised learning algorithm, but we leave the pursuit of this direction for future work.

\section{Conclusions}
\label{sec:conclusions}
In this work, we proposed a framework allowing for the granular generation of small, focused synthetic support sets to augment the performance of general backbone networks on real data classification tasks. Our framework exploits the wealth of information present in large pretrained models by controlling their output using reinforcement learning agents, so that optimal, explainable prompts can be generated over many training steps. Our framework produced excellent results on a variety of backbones, increasing classification accuracy by significant margins for no additional labelling or data cost.
{
    \small
    \bibliographystyle{ieeenat_fullname}
    \bibliography{main}
}


\end{document}